\documentclass{article}
\usepackage{ijcai20}

\usepackage{times}
\usepackage{soul}
\usepackage{url}
\usepackage[hidelinks]{hyperref}
\usepackage[utf8]{inputenc}
\usepackage[small]{caption}
\usepackage{graphicx}
\usepackage{amsmath}
\usepackage{amsthm}
\usepackage{booktabs}
\usepackage{algorithm}
\usepackage{algorithmic}
\usepackage{times}
\usepackage{epsfig}
\usepackage{amssymb}
\usepackage{multirow}
\usepackage{flushend}

\title{HandAugment: A Simple Data Augmentation Method for Depth-Based 3D Hand Pose Estimation}

\author{
Zhaohui Zhang$^1$
\and
Shipeng Xie$^1$\and
Mingxiu Chen$^{1}$\And
Haichao Zhu$^1$
\affiliations
$^1$Rokid Corportation Ltd.\\
\emails
\{zhaohui.zhang, shipeng.xie, cmxnono, haichao.zhu\}@rokid.com
}

\begin{document}

\maketitle

\begin{abstract}
Hand pose estimation from 3D depth images, has been explored widely using various kinds of techniques in the field of computer vision. Though, deep learning based method improve the performance greatly recently, however, this problem still remains unsolved due to lack of large datasets, like ImageNet or effective data synthesis methods. In this paper, we propose HandAugment, a method to synthesize image data to augment the training process of the neural networks. 
Our method has two main parts: 
First, We propose a scheme of two-stage neural networks. This scheme can make the neural networks focus on the hand regions and thus to improve the performance.
Second, we introduce a simple and effective method to synthesize data by combining real and synthetic image together in the image space. 
Finally, we show that our method achieves the first place in the task of depth-based 3D hand pose estimation in HANDS 2019 challenge.
\end{abstract}

\section{Introduction}
Hand pose estimation from a single depth image lays the foundation of human-computer interaction
technique on a head-mounted Augmented Reality (AR) device,
e.g., Microsoft Hololens, Magical Leap One. It has the advantage that users can provide input to devices efficiently. 
Despite recent remarkable progress, this problem still remains unsolved because of the large pose variation, large view point variation, self-similarities and self-occlusion of finger joints .

Recently, Deep Learning has become popular in the community of computer vision and also achieves state-of-the-art on the 3D hand pose estimation tasks. These methods can be roughly classified into two categories. The first category treats the input depth image as a single channel image and apply 2D convolutional neural network directly on the depth image. Representative methods are A2J~\cite{xiong2019a2j}, DeepPrior++~\cite{oberweger2017deepprior++}. The second category of methods use 3D information. These methods either convert depth images into 3D voxels~\cite{moon2018v2v}, ~\cite{ge20173d} or point clouds~\cite{ge2018hand} and then followed by 3D CNN or point net respectively.

\begin{figure}[t]
	\centering
	\includegraphics[width=1\columnwidth]{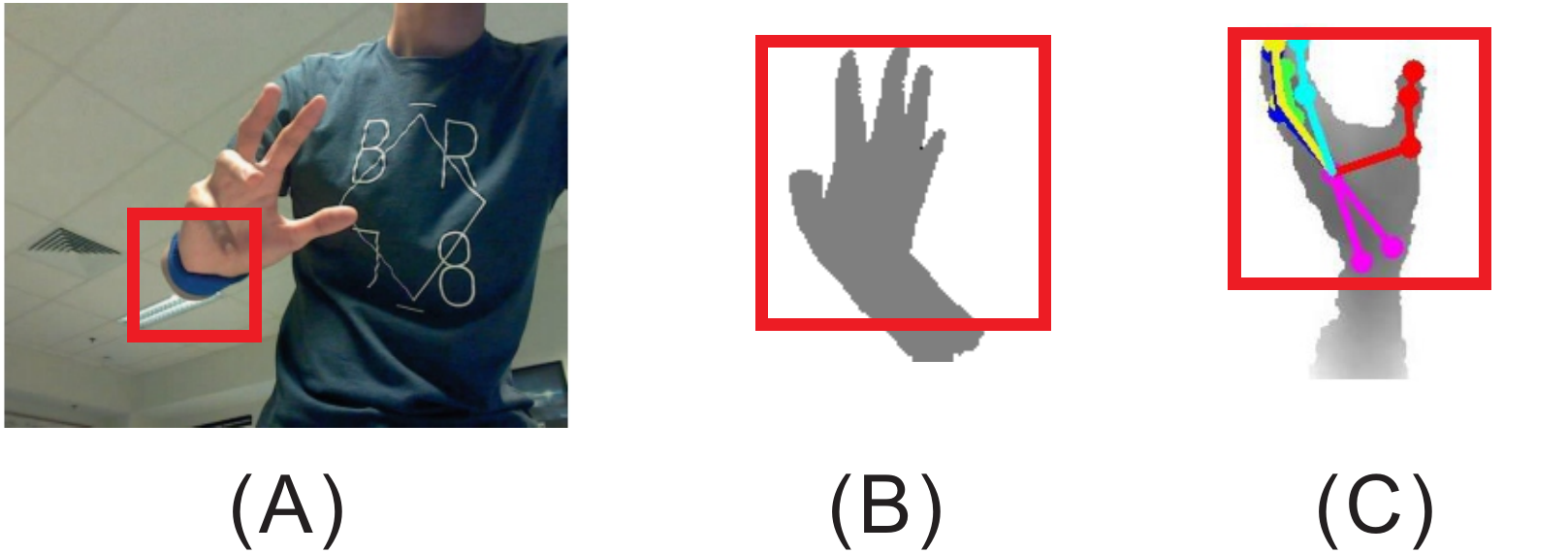}
	\caption{(A) A wrist band is detected to determine the hand region. (B) The hand region is estimated from input, however it can introduce arm or other foreground regions. (C) Using groundtrudth to extract hand region for training, however this is impractical in real application.}
	\label{fig:initialization} 
\vspace{-5mm}
\end{figure}

These neural networks are trained with hand regions extracted from depth images.
Intuitively, the quality of extracted hand region is important for hand pose estimation. However, the region extract method used in previous methods are naive.
For example, in~\cite{sinha2016deephand}, the users wears a colorful wristband which is used to determine the hand regions as show in Figure~\ref{fig:initialization} (A). This method is impractical in real cases.
In~\cite{chen2019pose}, hand regions are initialized using a shallow CNN. However, it can introduce arms or other foreground regions (Figure~\ref{fig:initialization} (B)). 
In~\cite{wan2018dense}, hand regions are obtained using groundtruth annotation which is not available in real application (Figure~\ref{fig:initialization} (C)).

In addition, these deep learning based methods are effective only if a large amount of training data is available. The data is usually collected and labelled manually, which is tedious and time consuming. This labeling problem is even worse for 3D computer vision problems which require to label with 3D data and this task is more difficult for humans. Recently, many works therefore focus on using computer graphics methods to synthesize image data~\cite{rad2018feature} and corresponding annotation data automatically. However, the resulting performances are usually suboptimal because synthetic images do not correspond exactly to real images. 

In this paper, we propose HandAugment, a method to synthesize image data to augment the training process of hand pose estimation neural networks. 
First, We propose a scheme of two-stage neural networks to tackle the hand region extraction. This scheme can gradually find the hand regions in the input depth maps.
Second, we propose a data synthesis method based on MANO~\cite{romero2017embodied}. Because synthetic images do not correspond exactly to real data, therefore we combine real data with synthetic images together.
Finally, we apply HandAugment to different datasets and the experiment shows our method can greatly improve the performance and achieves state-of-the-art results. Our method achieves the first place in the task of depth-based 3D hand pose estimation in HANDS 2019 challenge. Our codes are available upon request.

\section{Relate Work}
In this section we review related works of our proposed method, including depth-based 3D hand pose estimation and data augmentation method.

\subsection{Depth-Based 3D Hand Pose Estimation}
Hand pose estimation, has been explored widely using various kinds of techniques in the field of computer vision.
Related neural network-based hand pose estimation approaches using depth images are reviewed as follows.
The goal of hand pose estimation is to estimate the 3D location of hand joints from one or more frames recorded from a depth camera. 
The neural network based methods can be roughly classified into two categories: 2D and 3D deep learning respectively.
\paragraph{2D deep learning based approach.} 
The 2D deep learning based approaches estimate hand pose directly from depth images. Representative methods include a cascaded multistage method~\cite{chen2019pose}, a structure-aware regression approach~\cite{taylor2016efficient}, and 
hierarchical tree-like structured CNNs~\cite{madadi2017end}. Due to end-to-end working manner, deep learning technology holds strong fitting ability for visual pattern characterization. 2D CNN has already achieved great success for 2D pose estimation.
But these methods are unable to fully capture the 3D information from 3D hand poses, because these methods take depth maps as 2D single channel images for the input.

\paragraph{3D deep learning based approach.} 
To better reveal the 3D information within depth map for performance improvement. Some recent research tried 3D deep learning. The 3D deep learning based approach convert 2D depth images into 3D data structure, such as 3D voxel grids for~\cite{moon2018v2v} or D-TSDF volumes~\cite{ge20173d}. These 3D method is very accurate in 3D hand pose estimation problem and they produce state-of-the-art results. However, the 3D CNN is relatively hard to train due to a large number of parameters. Meanwhile, using 3D CNN also leads to high computational burden both on memory storage and running time. Therefore, it is less computational efficient than 2D methods.
Accordingly, HandAugment belongs to 2D deep learning based methods. We use a 2D CNN as the backbone network.

\subsection{Data Augmentation Methods}
Data augmentation is a strategy that can significantly increase the diversity of data for training deep network, without collecting additional training data. In recent years, Data augmentation techniques such as cropping, padding and flipping are commonly used in deep learning. This strategy can improve the performance of these data-driven tasks, suck like object recognition and hand pose estimation. It has already been widely used in recent work ~\cite{xiong2019a2j},~\cite{yang2019aligning} and ~\cite{oberweger2017deepprior++}. Most of these data augmentation methods use image transformation methods, including in-plain translation, rotating, scaling and mirroring. 
Specifically, for color-based methods, training images can be augmented by adjusting the hue channel of the color images~\cite{yang2019aligning}. For depth-based methods, images can be augmented by applying 3D transformation, 
such as ~\cite{ge20173d} which randomly rotates and stretches the 3D point cloud to synthesize training data.

Another way to synthesize training data is to use training samples rendered from 3D models ~\cite{hinterstoisser2018pre}. Such annotated samples are very easy to acquire due to the presence of large scale 3D model datasets. However, using synthetic data requires carefully designed train process to prevent the network from overfitting on the synthetic appearance of the data. This is due to the fact that the distribution of synthesize data is quite different from the distribution of real data.

Accordingly, our method use the rendered method to synthesize training data and perform data augmentation. We will show that the accuracy of hand pose estimation can be significantly improved by combining some real images and many synthetic images together through our proposed method.

\section{Method}
We first give an overview of HandAugment in Section \ref{sec:overview}. After that we present details about two stage network scheme in Section \ref{sec:net}. In Section \ref{sec:augment} we illustrate how to synthesize data for data augmentation. Finally, the implementation details are given in Section \ref{sec:detail}.

\begin{figure}[t]
	\centering
	\includegraphics[width=1\columnwidth]{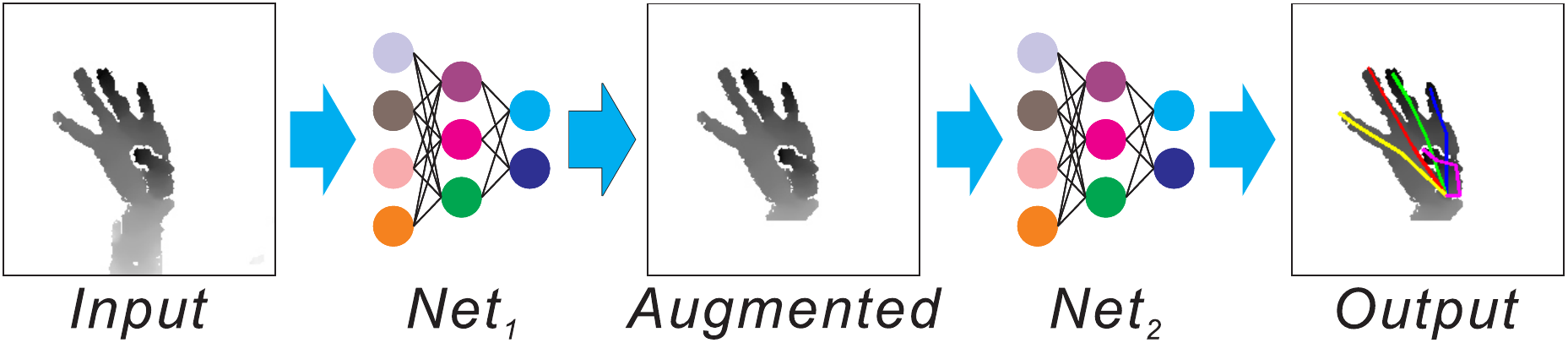}
	\caption{System Overview. The input depth image is feed into the first neural network $Net_1$ to obtain a augmented hand region. Then, this augmented hand region is feed into the second neural network $Net_2$ to estimate the hand pose.}
	\label{fig:overview} 
\end{figure}
\subsection{Overview}\label{sec:overview}
Given a depth image $I$, the task of hand pose estimation is to estimate the 3D locations $(x,y,z)$ of $N$ hand joints. We use a scheme of two-stage neural networks to estimate the hand poses, as illustrated in Figure~\ref{fig:overview}.
We first feed the input depth image into the first neural network (denoted as $Net_1$) which estimates an initial hand pose (denoted as $Pose_1$). 
Then, this initial hand pose $Pose_1$ is used to extract a augmented hand region from input depth image.
Finally, this augmented hand region is feed into the second neural network (denoted as $Net_2$) to estimate the final hand pose $Pose_2$. 


\begin{figure}[t]
	\centering
	\includegraphics[width=1\columnwidth]{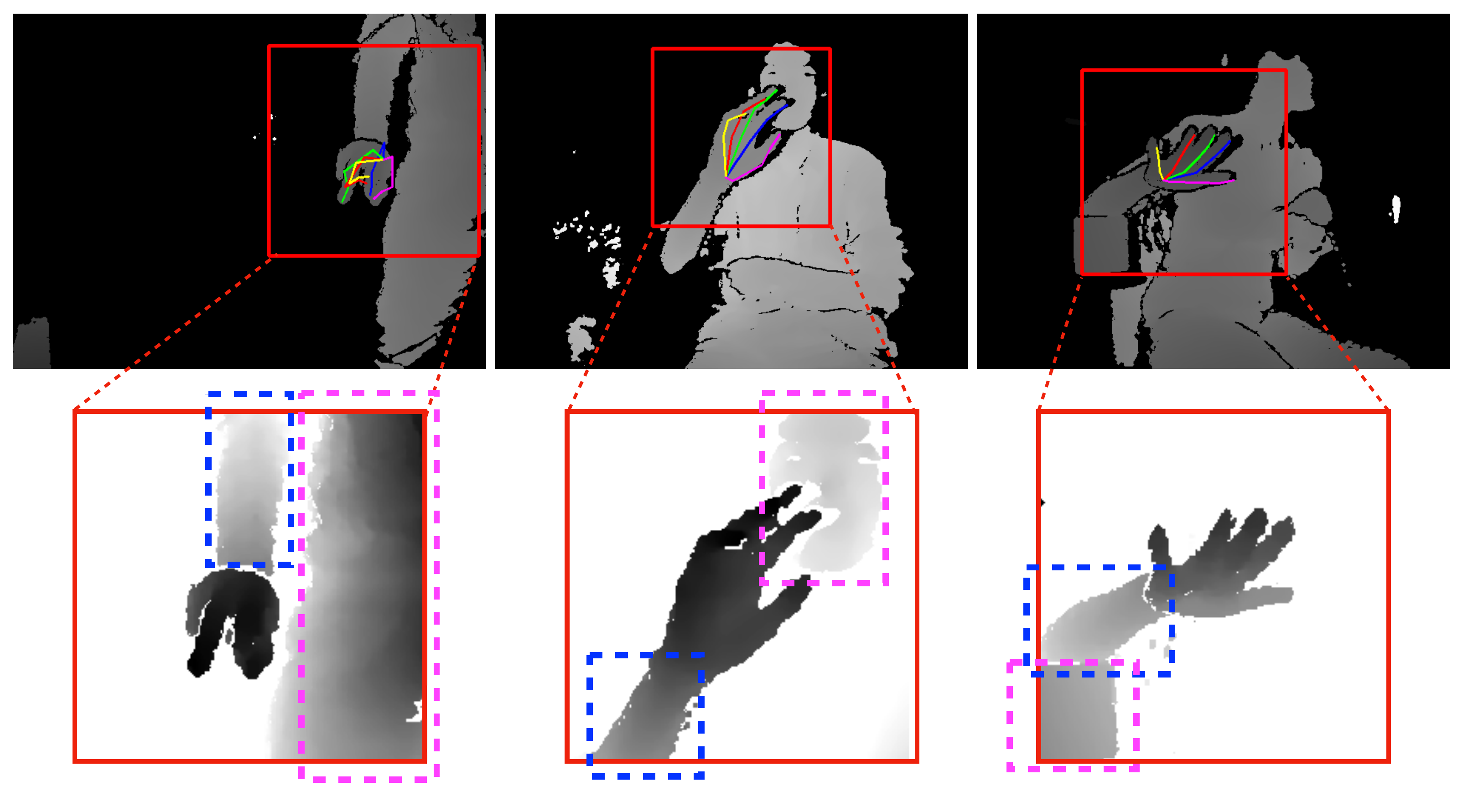}
	\caption{There is a lot of noise near the hand area in coarse patch $patch_1$, such as arm region (shown in blue dotted frame), human body and other background object (shown in magenta dotted frame).}
	\label{fig:noise} 
\end{figure}

\begin{figure}[t]
	\centering
	\includegraphics[width=1\columnwidth]{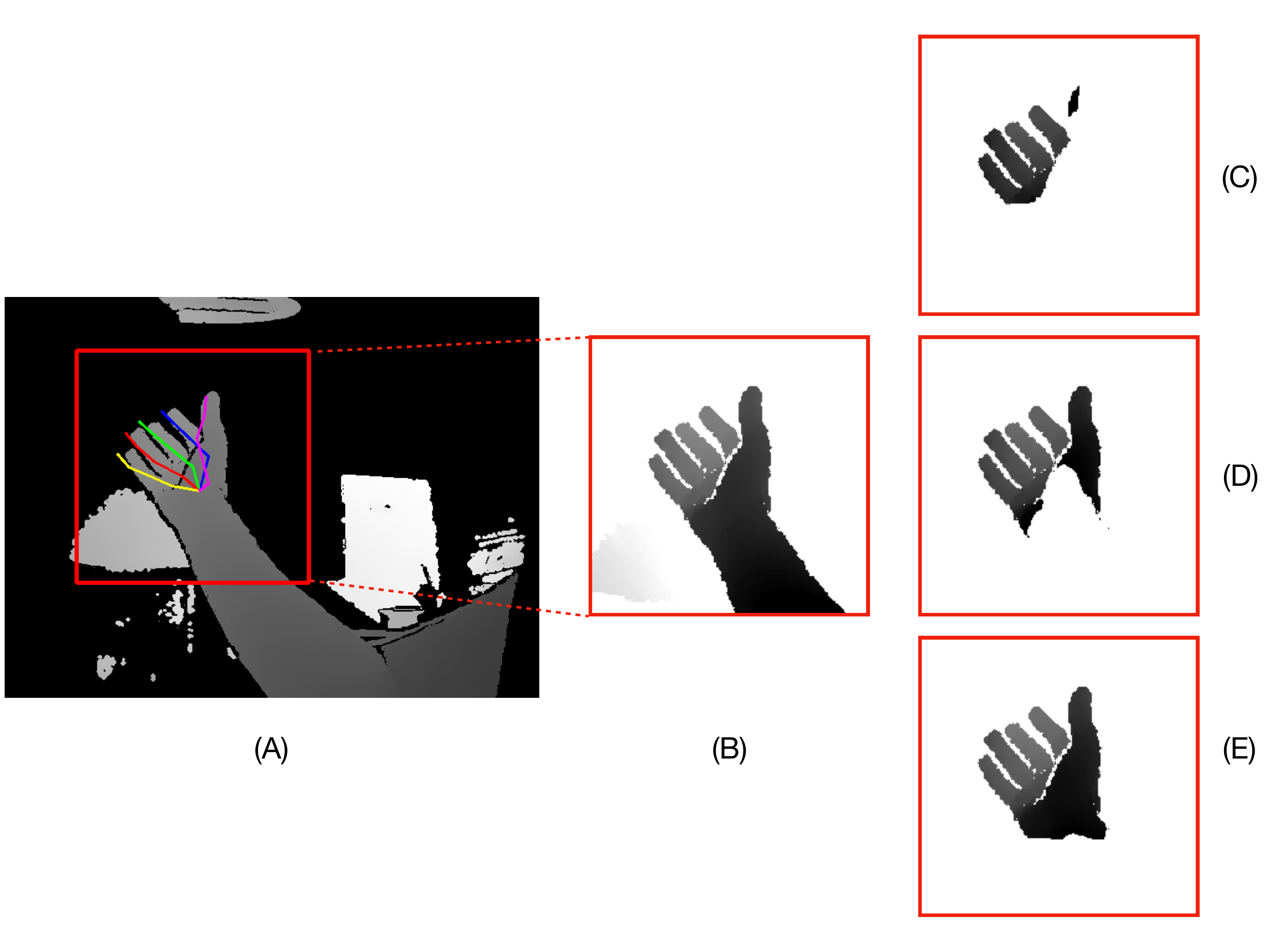}
	\caption{The influence of expanding range in region extraction. (A) The input depth image. (B) The coarse patch $patch_1$. (C) The augmented patch w/o range expanding and $z_{thickness}$ . (D) The augmented patch w/o $z_{thickness}$. (E) The augmented patch we proposed.}
	\label{fig:offset} 
\end{figure}

\subsection{Two-Stage Network Scheme}\label{sec:net}
We use a scheme of two-stage neural networks to estimate the hand poses. The input of the first stage neural network is a coarse patch extracted from input depth image.
This coarse patch, denoted as $patch_1$, usually contains noisy regions which appear around hand regions. These noisy regions can be arm regions or background objects, as shown in Figure~\ref{fig:noise}. Obviously, these noisy regions can degrade the performance. Thus, our solution is to remove these noisy regions from coarse patches to get augmented patches for hands.

To get the augmented hand patch, we first train the network $Net_1$ on coarse patch data $patch_1$ to predict an initial hand pose $Pose_1$.
After that, we find the maximum and minimum values from $Pose_1$ in 3D coordinate: $x_{min}$, $x_{max}$, $y_{min}$, $y_{max}$, $z_{min}$, $z_{max}$, and use them to determine a 3D bounding box.
Then, this 3D bounding box are used to loose crop from coarse patch and get an augment hand patch. Specifically, for any point $(x,y,z)$ in $patch_1$ that is out of the range $[x_{min}-x_{offset}, x_{max}+x_{offset}]$, $[y_{min}-y_{offset}, y_{max}+y_{offset}]$, and $[z_{min}-z_{offset}-z_{thickness}, z_{max}+z_{offset}]$ is removed from $patch_1$. While, $x_{offset}$, $y_{offset}$ and $z_{offset}$ are parameters to control extended range of 3D bounding box, $z_{thickness}$ denotes the thickness of finger.
The reason to extend the ranges of 3D bounding boxes is that the sizes of effective hand regions are usually larger than the sizes of hand skeleton bounding boxes. If using 3D bounding box directly without extending the range, the hand region extraction might not obtain full area of hand region in some cases, as shown in Figure~\ref{fig:offset}. Expanding range of bounding box can cover the gap between hand skeleton and hand skin (silhouette). In addition, due to the fact that the hand skin is always in front of hand skeleton in depth image, thus, we add parameter $z_{thickness}$ to handle this problem.

\begin{figure}[t]
	\centering
	\includegraphics[width=1\columnwidth]{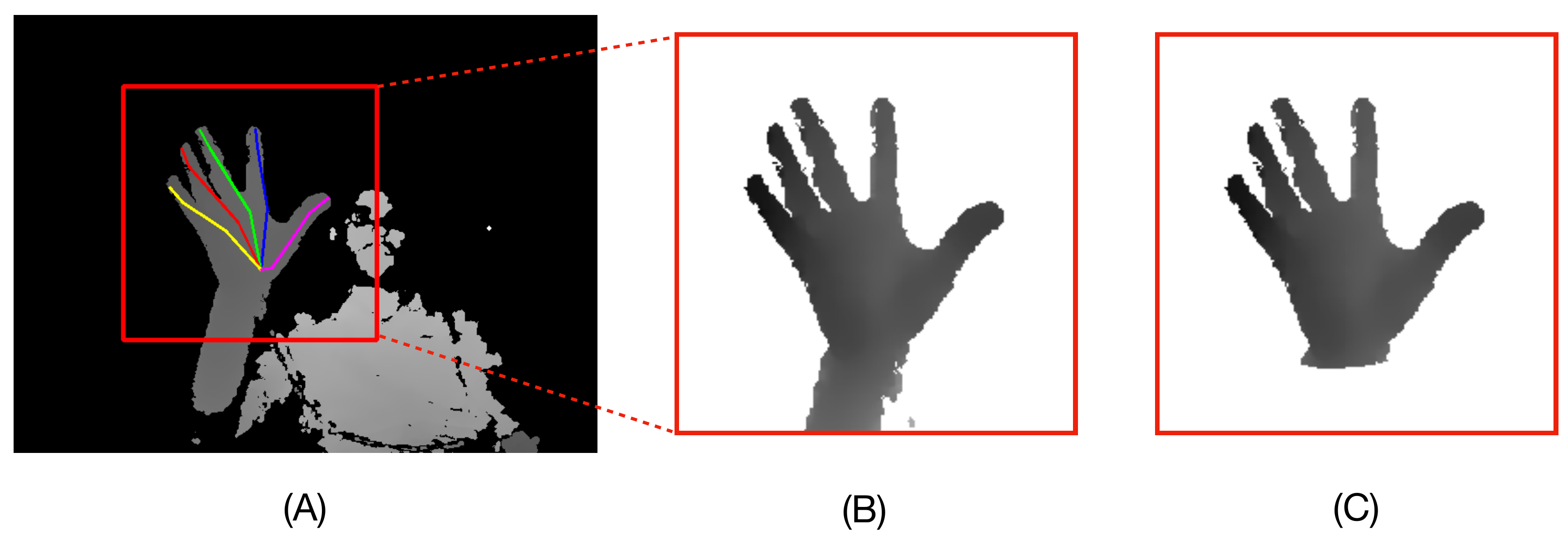}
	\caption{We use the pose estimate from $Net_1$ to augment the patch for $Net_2$. (A) The input depth image. (B) The input of $Net_1$. (C) The input of $Net_2$. }
	\label{fig:hand_clean} 
\end{figure}

The above procedure is denoted as first stage, and the augment hand patch obtained from first stage is denoted as $patch_2$.
In the second stage, the $patch_2$ is fed into $Net_2$ to get the final hand poses. This process is illustrated in Figure~\ref{fig:hand_clean}.

The architectures of our two networks are based on EfficientNet-B0 ~\cite{tan2019efficientnet}. We give the architecture of our modified EfficientNet-B0 in Table~\ref{tab:effcientnet_b0}. The input of these two networks are image patches cropped from input depth images. The cropped patches are resized to $224\times 224$ before feeding into the networks. The output of these two networks is a $3 \times 21$-dimensional vector indicates the 3D locations of the $21$ hand joints ($14$ hand joints for NYU experiment). Beside the input and output, the rest of the architectures are the same as that of the original EfficientNet-B0.
\begin{table} \scriptsize
\centering
\begin{tabular}{c|c|c|c|c}
\hline
Block & Operator &  Input Size     &  \#Channels & \#layers  \\ \hline
\hline
1 & Conv3x3 &  $224 \times 224$     &  32 & 1  \\ \hline
2 & MBConv1, k3x3 &  $112 \times 112 $     &  16 & 1  \\ \hline
3 & MBConv6, k3x3 &  $112 \times 112 $     &  24 & 2  \\ \hline
4 & MBConv6, k5x5 &  $56 \times 56 $     &  40 & 2  \\ \hline
5 & MBConv6, k3x3 &  $28 \times 28 $     &  80 & 3  \\ \hline
6 & MBConv6, k5x5 &  $14 \times 14 $     &  112 & 3  \\ \hline
7 & MBConv6, k5x5 &  $14 \times 14 $     &  192 & 4  \\ \hline
8 & MBConv6, k3x3 &  $7 \times 7 $     &  320 & 1  \\ \hline
9 & Conv1x1 \& Pooling \& FC &  $7 \times 7$     &  63 & 1  \\ 
\hline
\end{tabular}	\caption{\label{tab:effcientnet_b0} The summary of $Net_1$ and $Net_2$ architecture}
\end{table}

We train $Net_1$ and $Net_2$ with a Wing Loss~\cite{feng2018wing}, because the Wing Loss is robust for both small and large pose deviations. Given an estimated pose $p_i$ of the i-th joint and its corresponding ground truth $q_i$, the Wing Loss is defined as:
\begin{equation}
    L_{W}=\left\{
        \begin{array}{lcl}
            w\ln(1+\|v_i\|/\epsilon) & & \text{if }\|v_i\|<w \\
            \|v_i\|-C  & & \text{otherwise}
        \end{array}\right.
\end{equation}
where $v_i=p_i-q_i$, $w$ controls the width of non-linear part to be within $[-w, w]$, $\epsilon$ limits the curvature of the nonlinear part, and $C=w-w\ln(1+v_i/\epsilon)$ links the linear and non-linear parts together.

\subsection{Data Augmentation}\label{sec:augment}
\begin{figure}[t]
	\centering
	\includegraphics[width=1\columnwidth]{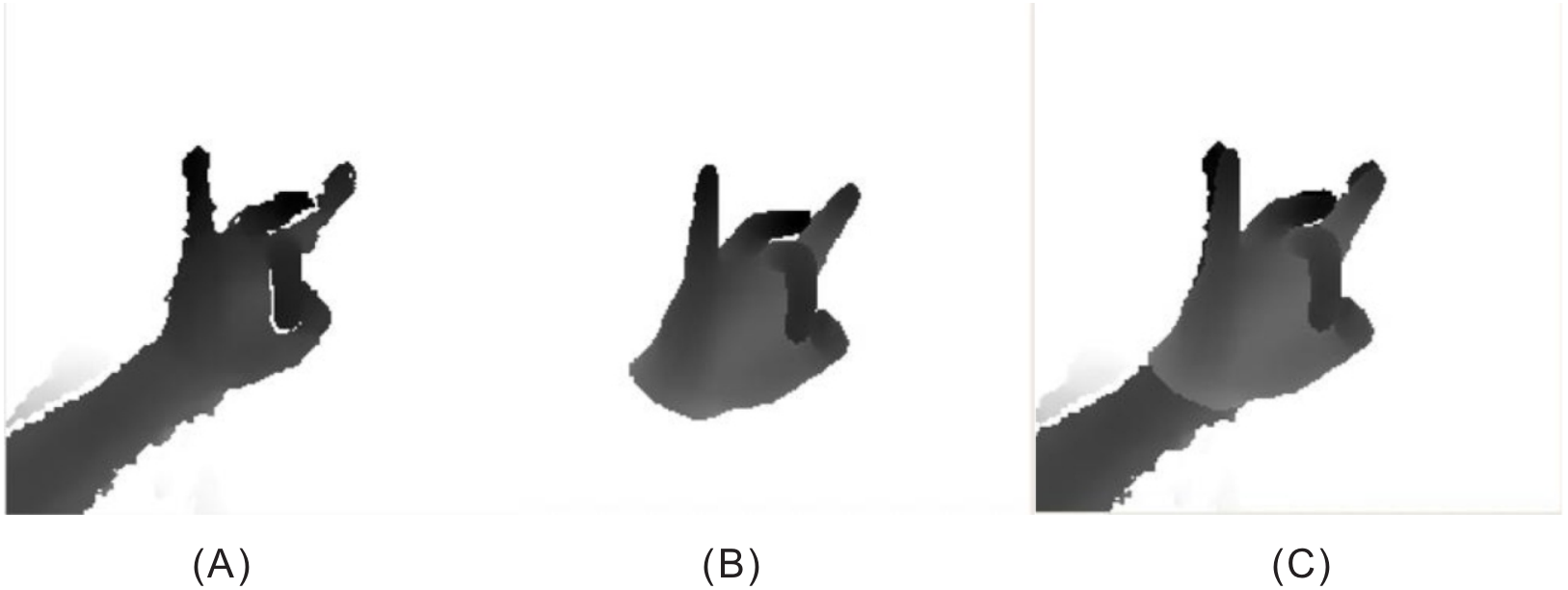}
	\caption{Data Synthesis. (A) A real image. (B) A synthetic image using corresponding MANO parameters. (C) The final synthetic image combining a real image and a synthetic image.}
	\label{fig:Synthetic} 
\end{figure}

Our method is based on MANO~\cite{romero2017embodied} to synthesize training data. MANO renders a depth image containing a right hand using three parameters: a camera parameter $c$, a hand pose parameter $a$ and a shape parameter $s$. The camera parameter $c$ is a 8-dimensional camera parameter including scale $c_s\in \mathbb{R}$, translation $c_s\in \mathbb{R}^3$ along three camera axes, and global rotation $c_q\in \mathbb{R}^4$ (in quaternion). The hand pose parameter $a$ is a 45-dimensional vector, and the shape parameter $s$ is a 10-dimensional vector.
To obtain MANO parameters, we use the HANDS19 dataset which uses gradient based optimization~\cite{baek2019pushing} to estimate MANO parameters from real images (Figure~\ref{fig:Synthetic} (A)). Then we use these estimated MANO parameters to synthesize images (Figure~\ref{fig:Synthetic} (B)). 

We have four strategies to prepare training data. 
The first two are to use real data and the original MANO parameters provided by HANDS19 directly. We do not add, remove or modify any data. These two dataset contain totally 170K images respectively. We call these two datasets as Real Dataset (RD) and Synthetic Dataset
(SD) respectively. 

The third strategy is to use a linear blending method to combine synthetic images with real images, that is because the distribution of SD and RD is different. An example is given in Figure~\ref{fig:Synthetic}. 
Given a synthetic image $I_s$ and its corresponding real image $I_r$, the final mixed image is given:
\begin{equation}
    I_{f}(i,j) = \begin{cases} I_s(i,j), & \mbox{if } I_s(i,j)>0 \\ I_r(i,j), & \mbox{if } I_s(i,j)=0 \end{cases}
\end{equation}
We create totally 170K mixed synthetic images. This dataset is denoted as the Mixed synthetic Dataset (MD).

Lastly, in order to generate more training data, we create new MANO parameters by add Gaussian noise to the original three MANO parameters $c$, $a$ and $s$ provided by HANDS 2019 dataset. The Gaussian distribution is obtained by assuming that each dimension of the three parameters are independent and the noise follow the same distribution as the original data. 
To generate the data, we add noise to only one of the three parameters (camera view, hand pose and shape parameters) or all of them. Totally, we create 400K images, 100K for camera view parameters, 100K for hand pose (articular) parameters, 100K for shape parameters and 100K for all of the three parameters. We denote this dataset as Noised
synthetic Dataset (ND).

\subsection{Implementation Details}\label{sec:detail}
\paragraph{Preprocessing.}
Similar to the previous method~\cite{chen2019pose}, we
extract a patch from the input depth image.
The patch center and patch size is determined by the metacarpophalangeal (MCP) joints of middle finger. Notice, we use a provided bounding box to get input patches on hands 2019 experiments where MMCP is not available. 
The patches are then resized to $224\times 224$.
The depth values of input patches are first truncated by the depth of MCP joint and then normalized into $[-1, 1]$.
These patches are then feed into neural networks.

\paragraph{Training.}
We train our two networks on a workstation equipped with a Intel Xeon Platinum 8160 CPU and two NVIDIA GEFORCE RTX 2080 Ti GPUs. We implement the networks using pytorch. To train $Net_1$, the batch size and learning rate are set 128 and $0.0006$ respectively and Adamax is used to optimize. 
A step-wise learning rate scheduler is used. The network is trained using all the training data, including RD, SD, MD and ND, and we have 640K images in total.
Then $Net_2$ is fine tuned from $Net_1$. The batch size and learning rate are also set 128 and $0.0006$ respectively. The optimizer is also Adamax. A step-wise learning rate scheduler is also used.
$Net_2$ is trained using RD and SD.
The parameter $w$ and $\epsilon$ from Wing Loss are empirically set as $100$ and $7.5$ respectively in all experiments.
All $x_{offset}$, $y_{offset}$, $z_{offset}$ are set to 30 mm, and $z_{thickness}$ is set to 20 mm.

\section{Experiment}
We first introduce datasets and evaluation metrics used in our experiments. Afterwards we compare our method with state-of-the-art methods. Finally we conduct extensive experiments for ablation study to discuss the effectiveness and robustness of different components of our proposed method.

\subsection{Datasets}\label{sec:datasets}

\par{\textbf{NYU Hand Pose Dataset~\cite{tompson2014real}.}}
The NYU hand pose dataset was collected using three Kinects from different views. The training set contains 72K images from 1 subject. And the test set contains 8.2K images from 2 subjects, while one of the subjects in test set doesn't appear in training set.
The annotation of 3D hand pose contains $36$ joints. Following the protocol of previous works~\cite{chen2019pose,guo2017region,moon2018v2v}, we only use images from the frontal view and pick $14$ of the $36$ joints for evaluation. Both annotations of training and test set are provided.

\par{\textbf{HANDS 2019 Dataset~\cite{hands2019}.}}
This dataset is sampled from BigHand2.2M~\cite{yuan2017bighand2}. The training set contains 175K images from $5$ different subjects. Some hand articulations and viewpoints are strategically excluded in the training set.
The test set contains 125K images from $10$ different subjects, $5$ subjects overlapping with the training set, exhaustive coverage of viewpoints and articulations. The annotations of hand poses contain $21$ joints, with $4$ joints for each finger and $1$ joint for the palm. The hand annotations are only available for the training set. Instead, the bounding boxes of the test set are provided. We use the HANDS 2019 official test tool to calculate test scores.
This dataset has large viewpoint, articulations and hand shape variations, which makes it a rather challenging dataset.

\begin{table} \scriptsize
\centering
\begin{tabular}{c|c|c|c|c|c}
\hline
Method & Main Error & I. & S. & A. & V. \\ \hline
BT~\cite{yang2019aligning} & 23.62 & 18.78 & 21.84 & 16.73 & 19.48  \\
IPR~\cite{sun2018integral} & 19.63 & 8.42 & 14.21 & 7.50 & 14.16  \\
V2V~\cite{moon2018v2v} & 13.76 & \textbf{3.93} & 11.75 & \textbf{3.65} & 7.50  \\
A2J~\cite{xiong2019a2j} & 13.74 & 6.33 & 11.23 & 6.05 & 8.78  \\
Ours & \textbf{13.66} & 4.10 & \textbf{10.27} & 4.74 & \textbf{7.44}  \\ 
\hline
\end{tabular}
\caption{Comparison Average joint 3D error(mm) and ranking result with state-of-art methods on HANDS 2019 dataset~\protect\cite{hands2019}. I., S., A. and V. stand for the errors of interpolation, shape, articulation and viewpoint, respectively. The main error is an extrapolation error on HANDS 2019 dataset. Details of the evaluation metric are described in section~\protect\ref{sec:metric}.}
\label{tab:hands19_result} 
\end{table}

\subsection{Evaluation Metric}\label{sec:metric}
 Average 3D joint error is average euclidean distance between predicted joint location and ground-truth for each joint over all test frames. We use the average 3D joint error as main evaluation metric in HANDS 2019 and NYU experiment.
 Furthermore, in HANDS 2019 dataset, there are five evaluation axes are calculated:
\begin{itemize}
\item[-] Total/Extrapolation: viewpoints, articulations and hand shapes not present in the training set. We
refer it as Extrapolation in the following.
\item[-] Articulation: articulations not present in the training set.
\item[-] Viewpoint: viewpoints not present in the training set.
\item[-] Shape: shapes not present in the training set.
\item[-] Interpolation: viewpoints, articulations and shapes present in the training set.
\end{itemize}


\subsection{Comparison with state-of-the-art methods}\label{sec:sota}

~\textbf{HANDS 2019 dataset}:
We compare our method with the state-of-the-art 3D hand pose estimation methods
. The results is listed in Table~\ref{tab:hands19_result}. It can be observed that:
\begin{itemize}
\item[-] On this challenging million-scale dataset, our method outperforms the other approaches in most of the score axes. This essentially verifies the superiority of our proposition.
\item[-] Our method reaches the lowest average joint 3D error in extrapolation, shape and viewpoint score axis, simultaneously. It demonstrates the robustness and generalization ability of our method.
\item[-] A2J and V2V are strong competitors to our method. And V2V even gets better score than our method in interpolation and articulation score axes. But their methods are the result of using carefully designed neural network architectures. As a consequence, it is much more complicated than our method which only uses a simple two-stage neural network. 
\end{itemize}

\begin{table}
\centering
\begin{tabular}{cc}
\hline
Method & Error (mm)  \\ \hline
DeepPrior++~\cite{oberweger2017deepprior++} & 12.24 \\
Pose-REN~\cite{chen2019pose} & 11.81 \\
HandPointNet~\cite{ge2018hand} & 10.50 \\
DenseReg~\cite{wan2018dense} & 10.20 \\
V2V~\cite{moon2018v2v} & 9.22 \\
A2J~\cite{xiong2019a2j} & \textbf{8.61}  \\
\hline
SS(our baseline) & 13.44 \\
TS (ours) & \textbf{9.02}  \\ 
\hline
\end{tabular}
\caption{Comparison Average 3D joint error with state-of-art methods on NYU dataset~\protect\cite{tompson2014real}.}
\label{tab:nyu_result} 
\end{table}

~\textbf{NYU Hand Pose dataset}:
Our method is compared with state-of-the-art 3D hand pose estimation methods.
The experiment result are given in Table~\ref{tab:nyu_result}. We can summarize that:
\begin{itemize}
\item[-] Our method is superior to the other methods in most cases. The exceptional case is that our method is slightly inferior to A2J method on NYU dataset. This is because NYU dataset provides ground truth annotations to extract hand regions for test data, but HANDS 2019 dataset only provides coarse bounding box to extract hand regions for test data. Therefore, using ground truth annotations for hand region extraction is more accurate than our proposed two-stage network scheme. However, it is impossible to get ground truth annotations in real application which makes A2J less practical than our method.
This demonstrates the robustness of HandAugment.
\item[-] Our proposed method decreases the error of baseline network (SS) from 13.44 mm to 9.02 mm with a $33\%$ improvement. This verifies the effectiveness of our proposed method.
\end{itemize}

\subsection{Ablation Study}\label{sec:ablation}
\subsubsection{Component Effectiveness Analysis}
The component effectiveness analysis within HandAugment is executed on HANDS 2019 dataset. We will investigate the effectiveness of two stage network scheme and our synthesized data strategy. 
Firstly, We build a baseline model $Model_0$ which is a single stage (SS) network and trained on real data (RD) only. Note that the single stage scheme only contains one EfficientNet-B0 network.
Secondly, We add two stage scheme and synthetic data to baseline model and denote them as $Model_1$ and $Model_2$, respectively.
Finally, We add both two-stage scheme and synthetic data to get our final model $Model_3$.
The result are given in Table~\ref{tab:ablation_all}.
It can be observed that:
\begin{itemize}
\item[-] The two-stage scheme remarkably improves the accuracy of hand pose estimation whether we use the data augmentation method or not. This verifies our observation that the extracted hand region is an important factor that affects the accuracy of predicted hand poses. Our proposed two-stage scheme can extract accurate hand regions for neural networks to estimate hand poses.
\item[-] Using synthetic data generated by our proposed method can tremendously decrease the average 3d joint errors of both the single-stage scheme and the two-stage scheme. This demonstrates the importance of synthetic data, and the effectiveness of our data synthesis method. 
\item[-] By combining all components together, our method finally gets $27.84\%$ improvement compared with the baseline model. This essentially verifies the effectiveness of HandAugment.
\end{itemize}

\begin{table}
\centering
\begin{tabular}{cccc}
\hline
Model Name & Network & Data & Error (mm) \\ \hline
$Model_0$ & SS & RD  & 18.93 \\ \hline
$Model_1$ & TS & RD & 16.50 \\ \hline
$Model_2$ & SS & RD+SD+MD+ND & 15.73 \\ \hline
$Model_3$ & TS & RD+SD+MD+ND & \textbf{13.66} \\
\hline
\end{tabular}	\caption{\label{tab:ablation_all} Experiments of different configuration of our method. SS and TS stand for the single stage and the two stage networks respectively. RD and SD stand for real data and synthetic data respectively.}
\end{table}

\begin{table}[t]
\centering
\begin{tabular}{ccc}
\hline
Model Name & Method & Error (mm) \\ \hline
$Model_0$ & ST & 18.93\\  \hline
$Model_{10}$ & TS$^*$  & 20.99 \\ \hline
$Model_1$ & TS$^\dagger$ & \textbf{16.50} \\
\hline
\end{tabular}	\caption{\label{tab:ablation_finetune} Experiments of different configuration of our method. * The two-stage scheme without fine-tuning on the second stage network. $\dagger$ The two-stage scheme with fine-tuning on the second stage network.}
\end{table}

\begin{table}[t]
\centering
\begin{tabular}{cccccc}
\hline
Model Name & Data & Error (mm) \\ \hline
$Model_{00}$ & SD & 39.66 \\ \hline
$Model_0$ & RD & 18.93 \\ \hline
$Model_{20}$ & RD+SD & 16.74 \\ \hline
$Model_{21}$ & RD+SD+MD & 16.22 \\ \hline 
$Model_2$ & RD+SD+MD+ND & \textbf{15.73} \\
\hline
\end{tabular}	\caption{\label{tab:systhetic_result} Effect of synthetic data. Notice that the single stage network is trained. All these results are obtained on a single stage network. RD and SD stand for real data and synthetic data respectively. MD and ND stand for the mixed synthesis data and the noised synthetic data respectively.}
\end{table}

\begin{table}[!t] \small
\centering
\begin{tabular}{ccc}
\hline
Model Name & Method & Error (mm) \\ \hline
$Model_{21}$ & No noise & 16.22 \\
$Model_{210}$ & Viewpoint & 15.94 \\
$Model_{211}$ & Articular & 15.82 \\
$Model_{212}$ & Shape & 16.13 \\
$Model_2$ & Viewpoint + Articulate + Shape & \textbf{15.73} \\
\hline
\end{tabular}	\caption{\label{tab:add_noise_result}  Adding noise to MANO parameters to generate synthetic data. All these results are obtained using the single stage scheme.}
\end{table}

\subsubsection{Effectiveness of The Fine-Tuning on The Second Stage Network}
Our two-stage scheme contains two neural networks and the second stage network is fine-tuned on the first stage network. To show how the fine-tuning can improve the performance, we give the results in Table~\ref{tab:ablation_finetune}. Note that these results are obtained using real data only. 
Our two-stage scheme without fine-tuning performs worse than the single-stage scheme. This is probably because the first stage network is over-fitting to the input hand regions of the first stage network. This over-fitting leads to the decreases of generalization.
The distribution of input data of the second stage is a subspace of the distribution of input data of the first stage. Obviously, it is easier to train a neural network in a subset if this network has been already trained on a super-set. Therefore, we fine tune the second stage network from the weights of the first stage network. The results show how it greatly improves the performance in Table~\ref{tab:ablation_finetune}. 

\subsubsection{Effectiveness of The Method for Synthesized Data}
We propose three strategies to synthesize training data as introduced in Section~\ref{sec:augment}. We train neural networks by using different combination of the three strategies and the results are given in Table~\ref{tab:systhetic_result}. Note that the results are obtained on the single-stage scheme. We can see that the performance is gradually improved as we add SD, MD and ND into training. This demonstrates that using our proposed strategies to synthesize training data fills the gap between the real data and synthetic data.

Furthermore, we show how different strategies of adding noise to synthesize data influence the performance. The results are listed in Table~\ref{tab:add_noise_result}. Note that these results are also obtained with the single-stage network. We add noise in one of the three parameters, including camera view point, hand articular pose and hand shape, or all of the three parameters. We can see that the performance is improved even we add noise into only one parameter.

\section{Conclusion}
In this paper, we propose HandAugment, a method to synthesize image data to augment the training of hand pose estimation method. 
First, We propose a scheme of two-stage neural networks to tackle the hand region extraction. This scheme can gradually find the hand regions in the input depth maps.
Second, we propose a data synthesis method based on MANO. We have three strategies to prepare the training data: using the original MANO parameters, mixed real and synthetic data and noised synthetic data.
Finally, we conduct several experiments to demonstrate that HandAugment is effective to improve the performance and achieves state-of-the-art results compared to existing method in Hands 2019 challenge. 

\newpage
\bibliographystyle{named}
\bibliography{egbib}

\end{document}